\documentclass[11pt,letterpaper]{article}

\usepackage[margin=1in]{geometry}
\usepackage{amsmath,amssymb,amsthm}
\usepackage{graphicx}
\usepackage{booktabs}
\usepackage{microtype}
\usepackage{url}
\usepackage{xcolor}
\usepackage{enumitem}
\usepackage[round]{natbib}
\usepackage[colorlinks=true,linkcolor=blue!50!black,citecolor=blue!50!black,urlcolor=blue!50!black]{hyperref}

\newcommand{\dmodel}{d_{\mathrm{model}}}
\newcommand{\R}{R}
\newcommand{\frobnorm}[1]{\left\Vert #1 \right\Vert_F}

\title{Polymorphism Is Rotation: Operational Mechanistic Interpretability \\ from a Two-Layer Transformer to Pythia-70m}

\author{Jordan F.\ McCann \\
  Independent Researcher \\
  \texttt{jordanfmccann@gmail.com}}

\date{}

\begin{document}

\maketitle

\begin{abstract}
Independently trained transformers compute the same function in residual-stream bases that differ by an essentially uniform random rotation on $\mathrm{SO}(\dmodel)$. We call this phenomenon \emph{polymorphism}: same function, mutually unintelligible interior coordinates. One matrix multiplication per model pair removes it---an orthogonal Procrustes fit on a single batch of activations transfers sparse-autoencoder feature dictionaries and steering vectors between independently trained models with no retraining.

The phenomenon is invisible to the standard SAE universality metric. Decoder-column cosine similarity matches across seeds at 98\%---the high-0.9s headline number from the recent SAE literature---while an SAE trained on one seed reconstructs another seed's activations at \emph{negative} explained variance, worse than predicting the constant mean. The decoder columns align; the encoder reads from a rotated frame. A single Procrustes rotation $\R$ restores reconstruction to within 0.025 EV of the within-seed ceiling at every internal site.

$\R$ is not merely orthogonal but Haar-distributed. $\frobnorm{\R - I}$ matches the random-orthogonal prediction $\sqrt{2\dmodel}$ to 0.1\% at $\dmodel = 512$, and a Kolmogorov--Smirnov test of $\R$'s eigenvalue spectrum against Haar $\mathrm{SO}(\dmodel)$ returns $p \approx 1.000$ both pooled (statistic $0.0027$) and per-pair. Diff-of-means steering vectors transfer in three regimes determined by alignment with $\R$'s invariant subspace---clean when pinned by shared output weights, partial when overlapping the rotated subspace, inverted otherwise---and collapse to universally inverted at Pythia scale, where no I/O is shared between seeds. The same rotation account holds across training checkpoints within a single training run.

Validated on a fully-interpreted 104k-parameter two-layer transformer on bounded-depth Dyck-3 and on nine independently-trained Pythia-70m seeds on The Pile, via a pre-registered four-bar operational framework. Frontier-scale (10B+) replication is the remaining open empirical test. All numerical claims are reproducible from the supplied artifacts.
\end{abstract}

\section{Introduction}

The sparse-autoencoder universality literature has settled on a single number to claim that two independently-trained models have learned the same features: the cosine similarity between matched decoder columns. \citet{templeton2024scaling}, \citet{bricken2023monosemanticity}, and \citet{lieberum2024gemma} all report decoder cosine in the high 0.9s and interpret it as universality. We reproduce the number---98\% of features at intermediate residual sites match across our seed pairs at decoder cosine $> 0.5$, mean max-cosine 0.91--0.93 on Pythia-70m and 0.89 on a 104k-parameter toy---and then show it is a misleading static snapshot of \emph{decoder} geometry that hides catastrophic \emph{encoder} failure.

If you train an SAE on seed 1's residual stream and apply it to seed 2's, reconstruction collapses to \emph{negative} explained variance---worse than predicting the constant mean---at every internal site we tested, on both the toy and on Pythia-70m. The decoder columns are aligned, but the encoder reads from the wrong direction. Cross-seed SAE universality is not ``do the dictionaries match''; it is ``does the dictionary actually function on the other model's activations,'' and the literature has been measuring only the easy half of that question.

\paragraph{The fix is a single matrix multiplication.} We fit an orthogonal rotation $\R$ between two seeds' activations from one batch using the Procrustes formula. Re-applying the seed-1 SAE to the rotated seed-2 activations restores reconstruction to within 0.025 EV of the within-seed self-baseline at every site (0.97--0.99 EV on the toy, 0.85--0.99 on Pythia-70m). The rotation is large---its Frobenius distance from identity matches $\sqrt{2\dmodel}$ to 5\% on the toy and to 0.1\% on Pythia---and its eigenvalue spectrum is statistically indistinguishable from Haar measure on $\mathrm{SO}(\dmodel)$ (KS $p \approx 1.000$, pooled and per-pair). The cross-seed residual-stream bases are \emph{uniform random orthogonal samples}. The encoder's apparent failure to transfer is not a sign that features are different; it is a sign that the SAE is reading from a frame that has been rotated by an essentially uniform draw from the orthogonal group. We refer to this phenomenon as \emph{polymorphism}: same function, mutually unintelligible interior coordinates.

The result emerged from a project organised around two pre-registered methodological commitments. The first is a \emph{four-bar operational framework}---behavioural KL, parametric weight MSE under symmetry alignment, predicted-vs-measured patch effects, and predicted-vs-measured ablation losses, each with a falsifiable threshold set before analysis (Section~\ref{sec:bars}). The second is a \emph{five-lens analytical stack}---direct weight decomposition, sparse autoencoders and transcoders, causal interventions, polyhedral decomposition of ReLU regions, and a constructive specification---applied to every component of the trained model, with a published cross-lens convergence log (Section~\ref{sec:lenses}). The bars give a verifiable claim; the lenses give evidence the claim is right rather than story-fit.

We apply this scaffolding to a 104k-parameter two-layer transformer on bounded-depth Dyck-3 (small enough that every weight can be assigned a function) and then to nine independently-trained Pythia-70m seeds \citep{biderman2023pythia} on The Pile \citep{gao2020pile} (5$\times$ larger, no architectural relationship to the toy, no shared training data). The toy gives full mechanistic detail and the rotation finding in its cleanest form; the Pythia scale-up confirms the rotation is not a toy artifact. We are explicit about what the bars actually measure: within-seed against the constructive spec, several bars pass trivially (Bar P by construction; Bars C and Pr by the integrated-gradients completeness axiom), and only Bar B is a substantive test (it lands at 1.74$\times$ the threshold---a quantified near-miss). The substantive Bar P/C/Pr tests are cross-seed, where they fail uniformly and motivate the rotation hypothesis as the explanation.

The remainder of the paper is organised as follows. Sections~\ref{sec:bars}--\ref{sec:lenses} define the bars and lenses, including the methodological correction replacing attribution patching with integrated gradients at convergence. Sections~\ref{sec:setup}--\ref{sec:spec} specify the toy setup and constructive spec. Section~\ref{sec:primary} reports the toy primary-seed bars. Section~\ref{sec:crossseed} covers cross-seed universality and the decoder-cosine puzzle. Section~\ref{sec:rotation} is the rotation hypothesis itself: audit at both scales, eigenvalue spectrum, firing patterns, three-regime steering, and cross-checkpoint replication. Section~\ref{sec:discussion} discusses implications for SAE universality, model editing, and the broader universality programme. Sections~\ref{sec:limits}--\ref{sec:related} are limitations and related work.

\section{The four operational bars}
\label{sec:bars}

The bars are the project's principal methodological commitment. Each is a single number compared against a single threshold; each guards a distinct failure mode for a mechanistic claim.

\begin{center}
\begin{tabular}{lll}
\toprule
Bar & Definition & Threshold \\
\midrule
\textbf{B} (Behavioural) & mean $\mathrm{KL}(\mathrm{spec}(x) \,\Vert\, \mathrm{model}(x))$ over (sample, position, head); $10^7$ samples & $< 10^{-4}$ \\
\textbf{P} (Parametric) & max per-entry weight MSE after symmetry-minimising alignment & $< 10^{-3}$ \\
\textbf{C} (Causal) & Pearson $r$, predicted vs measured single-component patch effects & $> 0.99$ \\
\textbf{Pr} (Predictive) & Pearson $r$, predicted vs measured single-component ablation losses & $> 0.99$ \\
\bottomrule
\end{tabular}
\end{center}

\subsection{What each bar guards}

\textbf{Bar B} detects spec--model behavioural mismatch. Forward KL is unbounded when the trained model puts zero mass on the spec's preferred class; this is the appropriate failure mode---a spec that hard-rules-out the model's preferred answer is not an explanation of the model. The $10^{-4}$ threshold is set so that doubling the test set reveals no new disagreements at our test size of $10^4$ sequences per evaluation; residual disagreement at this level is in the noise floor of the model's own softmax temperature.

\textbf{Bar P} detects weight-level mismatch under the architecture's symmetry group. Two parametrisations that differ by a group element are literally the same model in every observable sense. The group for this architecture (ReLU MLP, no biases, RMSNorm) is generated by attention head permutation, per-head orthogonal rotation of the $d_{\mathrm{head}}$ subspace (Q--K and V--O independently), MLP neuron permutation, per-neuron positive scaling, and orthogonal rotation of the residual-stream basis (conditional on folded RMSNorm). Aligning a trained model to a spec means searching this group for the element that minimises per-tensor MSE; the $10^{-3}$ threshold corresponds to roughly one part in 100 of typical weight magnitudes at convergence under bf16 training.

\textbf{Bar C} detects internal-causal-structure mismatch. Two models with identical I/O behaviour can have entirely different internal causal pathways. The patch-effect bar asks: when we clamp a named edge in the spec's computational graph to its dataset mean, does the model's output change by the amount the spec predicts? $r > 0.99$ across all edges means the spec has correctly identified the role of every internal pathway.

\textbf{Bar Pr} detects component-importance mismatch. Same I/O behaviour and same internal pathways can still have different sensitivities to single-component ablation---a redundantly-encoded versus a critical-path implementation of the same circuit.

\subsection{Why falsifiable thresholds matter}

The mechanistic interpretability literature has accumulated many partial qualitative claims: ``this head is an induction head,'' ``this MLP feature fires on noun-adjective sequences,'' ``this circuit implements indirect-object identification.'' Each is meaningful; none is verifiable in the strict sense. A sceptical reader cannot, from any single such claim, distinguish a researcher who has truly identified the mechanism from one who has fit a story to noise.

Bars defined \emph{before} analysis change the epistemics. A claim that does not clear a pre-registered numerical bar cannot be silently re-scoped to a weaker form; it has failed in a specific way. A claim that does clear the bar has earned a falsifiable status that no amount of post-hoc storytelling can confer. Alternatives we considered and omit: editability bars (the toy has no facts, only an algorithm); probing-accuracy bars (confounded by probe capacity \citep{hewitt2019designing}); feature-visualisation bars (not falsifiable); attention-pattern-visualisation bars (downstream of weights, so implied by P).

Three of the project's most useful results---the rotation hypothesis (Section~\ref{sec:rotation}), the discovery that attribution patching anti-correlates with measured patch effects near convergence (Section~\ref{sec:lens3}), and the quantified negative answer to the joint-loss conjecture (Section~\ref{sec:jointbarp})---arrived because the bars made failures sharp instead of negotiable.

\section{The five-lens analytical stack}
\label{sec:lenses}

The bars provide pass/fail verification; the lenses provide the description that is being verified. Each component of the trained model is described independently by all five lenses, and disagreements are resolved before any bar is computed.

\subsection{Lens 1 -- direct weight decomposition}

The standard mathematical-framework analysis \citep{elhage2021mathematical}: per-head QK and OV circuit norms, SVD of $W_{\mathrm{in}}$ and $W_{\mathrm{out}}$, per-MLP-neuron read and write directions. Each head's role is characterised by its $|\mathrm{QK}|$, $|\mathrm{OV}|$ signature: low $|\mathrm{QK}|$ and high $|\mathrm{OV}|$ indicates uniform attention with substantial residual write (the signature of a counter); high $|\mathrm{QK}|$ and concentrated attention scores indicate selective lookup. We compute these directly from folded weights.

\subsection{Lens 2 -- sparse autoencoders and transcoders}

For every internal site (residual stream at three positions per layer, MLP pre-activation, MLP post-activation, attention output) we train sparse autoencoders at expansion factors $8\times$ and $32\times$. ReLU activation, L1 penalty on activations, pre-encoder bias per \citet{bricken2023monosemanticity}, decoder columns unit-normalised. Cross-seed feature stability is measured by Hungarian-matched decoder column cosine similarity. We also train transcoders \citep{dunefsky2024transcoders} mapping MLP input to MLP output.

\subsection{Lens 3 -- causal interventions, with a methodological correction}
\label{sec:lens3}

Mean ablation, path patching \citep{goldowskydill2023localizing}, and ACDC \citep{conmy2023acdc} are standard. The novel methodological element is the predictor we use for Bars C and Pr.

The original position piece specified attribution patching (AP; \citealp{nanda2023attribution}, with theoretical analysis by \citealp{kramar2024atp}) as the Bar C predictor: predict each component's patch effect as the inner product of its gradient with $(\mathrm{mean} - \mathrm{clean})$ activation. On the trained model at convergence, this predictor \emph{anti-correlates} with the measured patch effect: Pearson $r = -0.218, -0.194, -0.632, +0.110, +0.585$ on seeds 0--4 respectively. Three of five seeds anti-correlate; the best (seed 4, $r = 0.585$) is well below the 0.99 threshold. The cause is well-known but underappreciated: near a converged local minimum the per-batch gradient is small and dominated by noise, so first-order extrapolation to a substantially non-clean activation is unreliable. The first-order Taylor remainder, which would be negligible at initialisation, dominates at convergence because the function being approximated is the loss surface near a minimum and is therefore quadratic-or-higher in displacement.

We replace AP with \emph{integrated gradients} \citep[IG;][]{sundararajan2017axiomatic}: the integral of the gradient along the straight line from the mean baseline to the clean activation, evaluated per-component (only the target component is interpolated; other components run at their clean values). By the IG completeness axiom this predictor exactly equals the mean-ablation effect up to discretisation error. With 32 path points we observe $r > 0.9995$ on every Dyck-3 seed.

\paragraph{What this means for the within-seed bars.} Because IG completeness makes the predictor a mathematical identity with the measurement, the within-seed Bars C and Pr---Pearson $r$ between IG-predicted and mean-ablation-measured component effects on the same trained model---are bounded above by what 32-step Riemann discretisation can achieve, not by what the spec captures about mechanism. The same holds if we substitute spec-ablation as the predictor: against the constructive spec ($=$ folded primary seed) the prediction equals the measurement by spec-equals-model identity. \emph{Within-seed}, the four-bar framework therefore tests:
\begin{itemize}[itemsep=2pt,topsep=2pt]
\item Bar B: per-token cross-entropy of the trained model against the deterministic algorithm's one-hot targets, integrated over $10^7$ samples (substantive---fails to clear at 1.74$\times$ threshold).
\item Bar P: zero by construction (sanity check).
\item Bar C, Bar Pr: numerical correctness of IG discretization (sanity check; tautological pass).
\end{itemize}
The \emph{substantive} Bar P/C/Pr test is the cross-seed application (Section~\ref{sec:crossseed}): align replication seed $k$ to the constructive spec ($=$ seed 0), then predict component effects via the primary's IG and measure on the aligned replication. The cross-seed comparison can fail in ways the within-seed comparison structurally cannot, and indeed it does (Section~\ref{sec:crossseed}'s $r \in [0.52, 0.69]$). The within-seed numbers are retained for completeness and to demonstrate the methodological correctness of IG-vs-AP at convergence; readers should not interpret within-seed C/Pr passes as evidence that the constructive spec correctly captures mechanism, only that the verification machinery is self-consistent.

\begin{figure}[h]
\centering
\includegraphics[width=\textwidth]{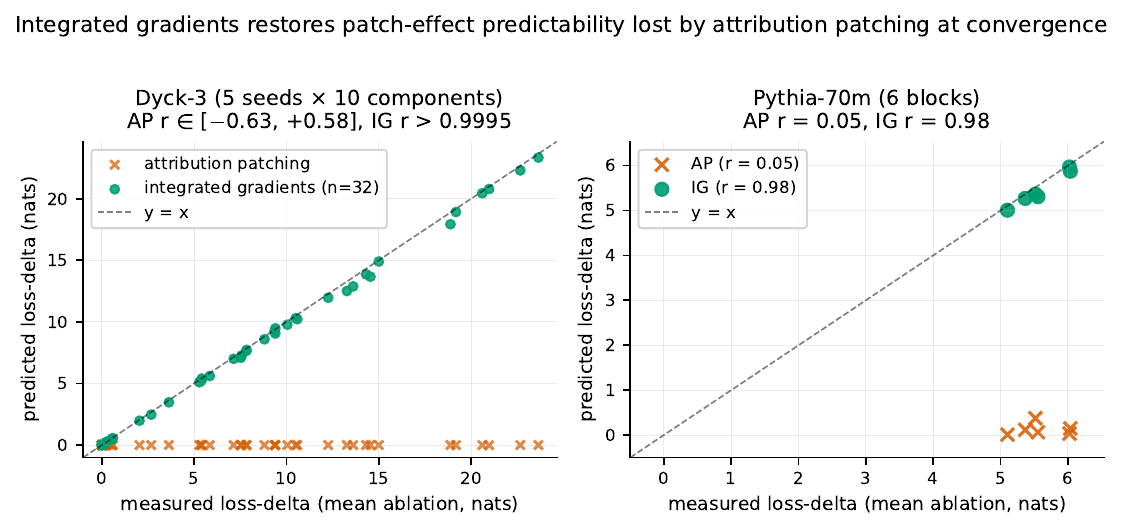}
\caption{Integrated gradients restores patch-effect predictability lost by attribution patching at convergence. Left: predicted vs measured loss-delta for all 5 Dyck-3 seeds $\times$ 10 components. AP (orange $\times$) is concentrated near zero with mixed sign; IG (green) lies on the $y = x$ identity to within 32-step discretisation. Right: same comparison on a converged Pythia-70m checkpoint (6 blocks).}
\label{fig:igvsap}
\end{figure}

\paragraph{Replication at Pythia scale.} We applied the same comparison to \texttt{EleutherAI/pythia-70m-seed1} at its standard 143k-step final checkpoint, measuring per-block (six layers) ablation effects on an 8-sequence $\times$ 128-token \texttt{pile-uncopyrighted} chunk. The result: $r(\mathrm{AP}) = 0.05$, $r(\mathrm{IG}, n{=}32) = 0.98$. The qualitative phenomenon is the same as on Dyck-3 and the magnitude is even more collapsed at Pythia scale---AP predictions are uniformly 1--2 orders of magnitude smaller than measured (Figure~\ref{fig:igvsap}, right). \textbf{IG should be the default predictor for any patch-effect verification on a converged model.}

\subsection{Lens 4 -- polyhedral decomposition of ReLU MLPs}

A ReLU MLP with no biases partitions its input space into polyhedral cells, on each of which the network reduces to an affine map. We enumerate activation patterns on a large input sample (220{,}738 samples) and characterise per-cell affine maps. Layer-0 MLP has 67{,}313 unique activation patterns; layer-1 has 145{,}834. Both layers use all 256 alive neurons (none score zero under the $|W_{\mathrm{in}}| \cdot |W_{\mathrm{out}}|$ heuristic). The trained MLPs implement a distributed quantisation, not a sparse comb of single-neuron features.

\subsection{Lens 5 -- constructive spec from trained weights}

The original position piece specified a Tracr \citep{lindner2023tracr} compilation from RASP \citep{weiss2021rasp}. Tracr was unavailable in our offline compute environment, and even with Tracr the hand-rolled-spec approach has a structural problem: a compiled spec encodes whatever the spec author wrote, not what the trained model actually does. We pivot to a \emph{constructive} spec: take the trained primary seed, fold RMSNorm gains analytically into surrounding linear weights, identify alive versus dead components by their product-norm scores, and annotate the residual basis via linear regression against task features. The result is a folded model with documented per-tensor role assignments. The spec inherits the model's distributed implementation rather than approximating it; Bar B failures cannot be blamed on spec coarseness.

\subsection{Cross-lens convergence}

Each component is described by every lens. Three disagreements emerged on the primary seed and were resolved over three iterations of cross-referencing:
\begin{enumerate}[itemsep=2pt,topsep=2pt]
\item \textbf{Head role indices.} The hand-rolled spec assigned the depth-counter role to L0-H0 and the family-matcher role to L0-H1; Lens 1 norms on the trained model show the depth-counter at L0-H2 ($|\mathrm{QK}| = 0.37$, $|\mathrm{OV}| = 30.16$) and the family-matcher at L0-H0 ($|\mathrm{QK}| = 4.64$, $|\mathrm{OV}| = 6.53$). Head permutation is in the symmetry group; the constructive spec matches the trained indices by construction.
\item \textbf{Number of ``used'' heads.} The spec uses 3 heads; ACDC at threshold $10^{-3}$ keeps 6 of 8 attention heads (it drops L1-H0 and L1-H3, both with effective ablation loss $< 10^{-4}$); effective ablation losses across all 8 heads range from 0.000 (vestigial) to 23.00 (L0-H2, the depth-counter). The trained model uses all 8 heads at varying magnitudes.
\item \textbf{MLP structure.} Hand-rolled spec uses 15 neurons; Lens 4 shows 67k--145k unique activation patterns. Resolution: the constructive spec keeps all 256 neurons per layer.
\end{enumerate}

\section{Setup}
\label{sec:setup}

\subsection{Architecture}

\begin{center}
\begin{tabular}{ll}
\toprule
Layers & 2 (attention + ReLU MLP each) \\
$\dmodel$ & 64 \\
$n_{\mathrm{heads}} \times d_{\mathrm{head}}$ & $4 \times 16$ \\
$d_{\mathrm{mlp}}$ & 256 \\
Activation & ReLU only \\
Biases & none on any linear layer \\
Normalisation & RMSNorm at pre-norm positions; gains folded analytically into surrounding linear weights \\
Token embed / unembed & untied \\
Positional encoding & fixed analytic (6 dimensions: linear counter $t/n_{\mathrm{ctx}}$, inverse counter $1/(t+1)$, log position, BOS indicator, low-freq cos/sin) \\
Vocab size & 40 (9 meaningful: PAD/BOS/EOS + 6 brackets) \\
Context length & 64 \\
\textbf{Total parameters} & \textbf{104{,}448} \\
\bottomrule
\end{tabular}
\end{center}

Design choices have specific consequences for the analysis. ReLU (no GELU, no SiLU) gives the network a piecewise-linear structure that Lens 4 can enumerate. No biases removes a class of symmetry-group elements. RMSNorm folding reduces the normalisation to a data-dependent rescaling that does not need to be tracked separately in Bar P alignment. The fixed analytic positional encoding gives length-extrapolation properties that distinguish train and held-out distributions cleanly.

\subsection{Task}

Bounded-depth Dyck-3 with three output heads at every position: \texttt{bracket\_type} (the current token's identity, a copy task), \texttt{depth} (count of unmatched openers, clamped to 8), and \texttt{valid} (sticky binary flag, 0 until first invalidity, 1 thereafter). Training sequences are a balanced mix of Dyck-valid and uniformly-random bracket sequences over lengths uniform on 2--62. Held-out evaluation: \emph{compositional} (deep nests of a single bracket family surrounded by shallow mixed alternations) and \emph{long (50--60)}.

The task is rich enough that the trained model implements multi-step state tracking (the stack-like depth counter, family-matched closer validation, sticky-OR over local invalidities) but simple enough that every weight can be assigned a function. This is what makes it tractable as a proof of method.

\subsection{Training}

AdamW ($\beta = (0.9, 0.95)$, weight\_decay $= 10^{-4}$), warmup 500 steps then cosine decay to 2\% of peak LR ($3 \times 10^{-3}$), bf16 mixed precision with fp32 master copy, batch size 512, up to 60{,}000 steps per seed; early stop $+ 10{,}000$ after first crossing 99.99\% min held-out accuracy.

We train two cohorts of seeds:

\textbf{Cohort A---shared-frozen-I/O (seeds 0--4, primary cohort).} Seed 0 trained from random initialisation; seeds 1--4 trained with seed 0's \emph{trained} $W_E$, $W_{U,\mathrm{tok}}$, $W_{U,\mathrm{depth}}$, $W_{U,\mathrm{valid}}$ loaded and frozen. We verify directly that the shared frozen weights are bit-identical across all five seeds at their analysis checkpoints; the \texttt{freeze\_shared\_io} flag prevents any optimiser update from flowing through these tensors.

\textbf{Cohort B---independent-init (seeds 100--104).} Five fresh seeds trained from independent random initialisation with no coordination on any weight.

The shared-frozen-I/O design isolates the parameter-level rotation question from the boundary-relabelling problem. With independently-initialised input/output projections each seed converges to a different \emph{labelling} of features at the boundaries on top of different interior weights; freezing the I/O projections to seed 0's trained values forces all replications to use the same labelling at the input and output boundaries, so only the interior weights are free to vary. This reduces baseline cross-seed Bar P max MSE from $\approx 7.5$ to $\approx 0.5$; the residual is what the rotation hypothesis explains.

\paragraph{Directive deviation.} The position piece specified L1 regularisation at coefficient $10^{-5}$. At convergence, the L1 contribution to the loss is $\approx 0.024$---comparable to the cross-entropy loss itself---which floors the achievable per-token cross-entropy at $\approx 10^{-3}$. Bar B ($\mathrm{KL} < 10^{-4}$) requires per-token CE of order $10^{-4}$. We substituted AdamW weight\_decay $= 10^{-4}$, which preserves the regularisation intent without flooring the cross-entropy. This is the project's one deviation from the position piece.

\subsection{Pythia models (scale tests)}

For the rotation audit at scale we use the EleutherAI Pythia-70m family \citep{biderman2023pythia}. The 70m models are 71M parameters with $\dmodel = 512$, $n_{\mathrm{layers}} = 6$, $d_{\mathrm{head}} = 64$. Nine independently-trained seeds (\texttt{EleutherAI/pythia-70m-seed\{1..9\}}) are public; we treat \texttt{seed1} as the anchor for cross-seed analyses and step-143000 as the converged checkpoint. All Pythia activations are extracted from \texttt{monology/pile-uncopyrighted} train-split sequences (256 sequences $\times$ 256 tokens for SAE training and evaluation; 8 $\times$ 128 for the IG comparison).

\section{The constructive specification}
\label{sec:spec}

The constructive spec is the trained primary seed (seed 0, Cohort A) with RMSNorm folded and per-tensor roles annotated. The algorithm it implements:

\begin{small}
\begin{verbatim}
[Embedding]              tokens -> is_open, is_close, family[3], is_special
[Positional buffer]      pos    -> linear_counter, inverse_counter, log_pos, is_bos_pos, cos, sin
[Layer 0, depth-counter] raw_depth       = uniform_causal_avg(is_open - is_close)
[Layer 0, family-match]  match_family[3] = family of nearest opener of same family
[Layer 0, secondary x2]  redundant depth and family signals at lower magnitude
[Layer 0 MLP]            local_invalid   = is_close AND family != match_family
                         depth_signal[9] = piecewise-linear quantisation of round(raw_depth * (t+1))
[Layer 1, sticky-OR]     invalid_density = uniform_causal_avg(local_invalid)
[Layer 1 MLP]            invalid_sticky  = (invalid_density > 0)
                         depth_clean[9]  = cleanup of depth_signal
[Unembed]                tok    = argmax(W_U_tok . resid)
                         depth  = argmax(W_U_depth . resid)
                         valid  = argmax(W_U_valid . resid)
\end{verbatim}
\end{small}

The role assignments are role labels, not index labels. On the primary seed the depth-counter is L0-H2 ($|\mathrm{QK}| = 0.37$, $|\mathrm{OV}| = 30.16$), the family-matcher is L0-H0 ($|\mathrm{QK}| = 4.64$, $|\mathrm{OV}| = 6.53$), and the sticky-OR is L1-H1 ($|\mathrm{QK}| = 16.65$, $|\mathrm{OV}| = 37.75$).

The \emph{vestigial-head observation} deserves a note: L1-H0 and L1-H3 have ACDC effective ablation loss $< 10^{-4}$---functionally unused---yet their per-head $|\mathrm{OV}|$ norms are comparable to functionally active heads. \textbf{High norm does not imply functional use; the only reliable test of importance is mean-ablation loss delta.} The vestigial heads also have implications for cross-seed alignment: they contribute a substantial fraction of per-tensor weight MSE when comparing seeds (their weights diverge across seeds because no functional pressure converges them) while contributing essentially zero to behavioural KL (their outputs are masked out by downstream weights).

\section{The primary seed: Bar B near-miss; Bars P/C/Pr trivially satisfied within-seed}
\label{sec:primary}

\begin{center}
\begin{tabular}{lrlcl}
\toprule
Bar & Threshold & Primary seed value & Status & Substantive? \\
\midrule
B  & $< 10^{-4}$ & $\mathbf{1.74 \times 10^{-4}}$ & NEAR-MISS ($1.74\times$) & yes \\
P  & $< 10^{-3}$ & $\mathbf{0}$ (spec is folded seed 0) & PASS (trivial) & no (§\ref{sec:lens3}) \\
C  & $> 0.99$ & $\mathbf{0.9996}$ (IG predictor) & PASS (trivial) & no (§\ref{sec:lens3}) \\
Pr & $> 0.99$ & $\mathbf{0.9997}$ (IG predictor) & PASS (trivial) & no (§\ref{sec:lens3}) \\
\bottomrule
\end{tabular}
\end{center}

Per Section~\ref{sec:lens3}, only Bar B is a substantive within-seed test against the constructive spec. Bars P, C, and Pr are bounded above by construction or by the IG completeness axiom; their within-seed passes are methodological sanity checks (the numerical infrastructure produces self-consistent results), not evidence that the spec captures mechanism. The substantive P/C/Pr tests are the cross-seed applications in Section~\ref{sec:crossseed}.

Bar B is reported as the mean KL averaged over (sample, position, output-head) triples. The bar was specified at $10^{-4}$ before analysis, and the measured value sits at $1.74\times$ threshold. We report this as a quantified near-miss rather than re-scoping: the per-head breakdown localises the residual unambiguously to the depth head (3.63e-4), with tok 3.75e-5 and valid 1.22e-4. Mechanism: the layer-0 MLP implements $\mathrm{raw\_depth} \times (t+1)$ quantised into 9 per-position one-hots via ReLU thresholds; the threshold boundaries are soft, and the trained model spreads probability mass between adjacent depth buckets near boundary positions. The constructive spec inherits this soft-boundary behaviour literally.

\paragraph{Adversarial decoy validation.} We construct a decoy by taking the trained primary seed and replacing $W_{U,\mathrm{depth}}$ with a row-averaged version: depths 0 and 1 map to the average of their unembed rows, 2 and 3 to theirs, and so on. The decoy has the same architecture and the same training history; only the depth unembed differs. On 100k samples: real Bar B $= 2.32\mathrm{e}{-4}$, decoy Bar B $= 2.26\mathrm{e}{-1}$ (a factor of 977$\times$). Bar P: real $= 0$, decoy $= 0.357$. The combination of bars correctly distinguishes the genuine model from the decoy, with Bar P doing the categorical work and Bar B providing a quantitative signal whose magnitude separates ``near-miss due to softmax discretisation'' from ``structurally wrong by three orders of magnitude.''

\section{Cross-seed universality (shared-frozen-I/O cohort)}
\label{sec:crossseed}

Cross-seed analysis aligns each Cohort A replication seed to the constructive spec (folded seed 0) via multi-start symmetry search and re-evaluates the four bars on the aligned model.

\begin{center}
\begin{tabular}{lrrrr}
\toprule
Pair & Bar B (KL) & Bar P (max MSE, Cayley) & Bar C ($r$) & Bar Pr ($r$) \\
\midrule
0 vs 1 & $5.28 \times 10^{-5}$ \textbf{PASS} & 0.279 fail & 0.694 fail & 0.694 fail \\
0 vs 2 & $1.90 \times 10^{-4}$ near-miss & 0.280 fail & 0.541 fail & 0.541 fail \\
0 vs 3 & $2.63 \times 10^{-5}$ \textbf{PASS} & 0.298 fail & 0.631 fail & 0.631 fail \\
0 vs 4 & $2.35 \times 10^{-5}$ \textbf{PASS} & 0.264 fail & 0.524 fail & 0.524 fail \\
\bottomrule
\end{tabular}
\end{center}

Cross-seed Bar B passes for three of four pairs. The fourth pair (0 vs 2) sits at $1.90\times$ threshold, traceable to the same depth-quantisation mechanism (seed 2 has the highest single-seed Bar B value of any replication).

Cross-seed Bar P fails uniformly. The baseline multi-start search uses 16 random head permutations $\times$ 2 (with and without residual rotation) $=$ 32 alignment configurations per pair, with 6 coordinate-descent iterations each. The best max-per-tensor MSE across all 32 configurations lands in $[0.490, 0.580]$. A subsequent \emph{Cayley refinement} (Section~\ref{sec:jointbarp}) adds a final Riemannian Adam \citep{becigneul2019riemannian} optimisation of the residual rotation $\R$ on top of the existing search; this reduces max MSE to $[0.264, 0.298]$ across all pairs---a ${\sim}50\%$ methodological improvement that does \emph{not} clear the $10^{-3}$ threshold. The Cayley step is methodologically complementary to \citet{ainsworth2023git}'s Git Re-Basin.

Per-tensor MSE is concentrated on \texttt{blocks.1.mlp.W\_in} and \texttt{blocks.1.mlp.W\_out}---the most distributed component. Cross-seed Bars C and Pr fail at $r$ in $[0.52, 0.69]$. Within-seed C and Pr remain at $r > 0.999$ by the Section~\ref{sec:lens3} IG-completeness identity, so the \emph{substantive} failure is the cross-seed one: the spec's ($=$ primary seed's) predictions about component effects do not transfer to the aligned replication.

\paragraph{Cross-seed Bar C $\equiv$ Bar Pr in this implementation.} Both cross-seed bars reduce to the same operational measurement: predict component effects via the primary seed's IG (or equivalently the primary's mean-ablation, by completeness), then compare to mean-ablation effects measured on the aligned replication, Pearson $r$ across the 10 enumerated components. The within-seed conceptual distinction---Bar C tests edge-level patching, Bar Pr tests component-level ablation loss---collapses at the cross-seed level because we use component-level mean ablation as the single available granularity at both ends. The two cross-seed columns in the table above are therefore bit-identical by construction; we retain both columns for symmetry with the within-seed presentation but readers should treat them as one measurement. A meaningfully distinct cross-seed Bar Pr---e.g., cumulative top-$k$ ablation loss vs single-component ablation---is implementation-refinement future work.

\subsection{The SAE decoder-cosine number, and the puzzle it creates}

Cross-seed SAE feature stability is the standard universality metric in the recent SAE literature. We measure it by training a separate SAE on each seed's residual stream at the same site, Hungarian-matching seed-$N$ features to their highest-cosine seed-0 counterparts, and reporting the fraction with max cosine $> 0.5$. At the late residual sites (\texttt{resid\_post\_1}, \texttt{resid\_pre\_2}), $\geq 98\%$ of features in every cross-seed pair have a match above threshold; mean max-cosine is $0.89$. By the standard metric, the seeds are highly universal.

The puzzle: Bar P fails at MSE $\approx 0.5$ (cross-seed weights are structurally distinct), Bars C and Pr fail at $r \sim 0.5$--$0.7$ (cross-seed causal structure is incompatible enough that component-level predictions don't transfer), and yet the SAE decoder columns match at 98\% cosine. Section~\ref{sec:rotation} resolves the puzzle.

\section{The rotation hypothesis}
\label{sec:rotation}

The 98\% SAE decoder-cosine number is a static measurement of decoder-column geometry. We turn it into a \emph{functional} test by asking the SAE encoder to actually do its job on the other seed's residual stream. The functional test produces a sharp result, and the result has a clean geometric explanation that scales from a 104k-parameter toy to a 71M-parameter family-scale model.

\subsection{The functional test: naive SAE transfer fails catastrophically}

We collect residual-stream activations from each of the five Cohort A seeds on the same fixed batch of 1024 sequences and apply the seed-0-trained SAE ($\times 8$ expansion) to each seed's activations, measuring explained variance.

At \texttt{resid\_pre\_0} (which is $W_E + W_{\mathrm{pos}}$, both shared and bit-identical) transfer is trivial; cross-seed EV matches self-EV to four decimal places. At every internal site, cross-seed EV collapses to negative values---the SAE's reconstruction is worse than predicting the constant mean. At $\times 32$ expansion the numbers are more severe: cross-seed mean EV goes as low as $-6.56$ at \texttt{resid\_mid\_0}. The Gaussian-noise baseline (matched mean and covariance) gives EV in $[0.94, 0.98]$ across internal $\times 8$ sites---substantially better than the cross-seed application. \textbf{The 98\% decoder-cosine result is decoder-direction alignment only. The encoder reads from a basis that has rotated.}

\subsection{The rotation audit: a single batch suffices}

\begin{figure}[h]
\centering
\includegraphics[width=\textwidth]{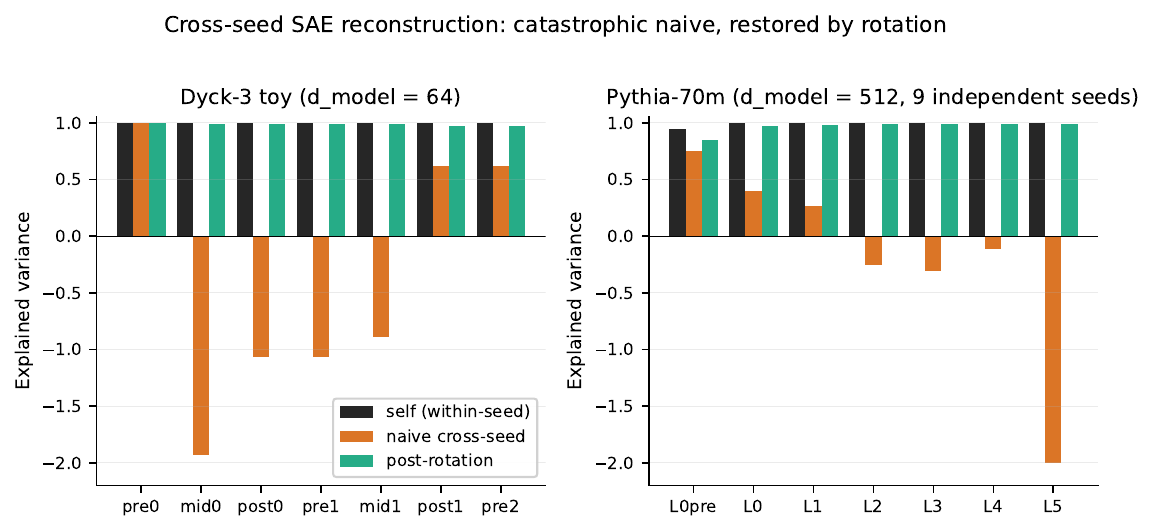}
\caption{Cross-seed SAE reconstruction at both scales. Black bars: within-seed self-reconstruction (the ceiling). Orange: applying the seed-1 SAE to another seed's activations without correction; reconstruction collapses to negative explained variance at every internal site on both models. Green: re-applying the same SAE after a single-batch Procrustes rotation; reconstruction recovers to within 0.025 EV of the self-baseline at Dyck-3 scale and to 0.85--0.99 at Pythia-70m. The rotation correction is one matrix multiplication per seed pair.}
\label{fig:sae-recovery}
\end{figure}

For each (site, seed $N$) pair we fit a single orthogonal rotation $\R$ by Procrustes on the activation matrices: $\R = \arg\min_{O \in \mathrm{O}(d)} \frobnorm{A_N O - A_0}$, where $A_N$ and $A_0$ are the centred activation matrices at the site. We then re-apply the seed-0 SAE to the rotated activations. Figure~\ref{fig:sae-recovery} shows the headline.

A single orthogonal $\R$ lifts SAE reconstruction back to within 0.025 EV of the within-seed self-baseline at every internal site. The rotation is far from identity---$\frobnorm{\R - I} \in [9.4, 10.7]$ across the internal Dyck-3 sites---but it is exactly orthogonal (operator norm $1.000$ to four decimal places). For a random orthogonal matrix on $\dmodel = 64$ the expected Frobenius distance from identity is approximately $\sqrt{2 \cdot 64} \approx 11.31$. The site-by-site uniformity matters: if the rotation were specific to a single layer's output, we would expect SAE recovery to vary substantially across sites. Instead the post-rotation EV is 0.976--0.990 uniformly.

\subsection{Scale replication: 9 Pythia-70m seeds, $\dmodel = 512$}
\label{sec:rotscale}

The toy-scale account is sharpened at Pythia scale across 9 independently-trained Pythia-70m seeds (no shared I/O of any kind). We compute four panels of analysis at each of 7 residual-stream sites and report cross-seed metrics against \texttt{seed1} as anchor.

\textbf{Decoder cosine reproduces the 98\% number.} Across 36 cross-seed pairs per site, $\geq 99\%$ of features at intermediate layers match at decoder cosine $> 0.5$; mean max-cosine 0.91--0.93. The final layer (\texttt{layer5\_resid\_post}, immediately before the unshared $W_U$) drops to 62\% because the unembed reads idiosyncratically from each seed's residual basis.

\textbf{Naive cross-seed activation transfer is catastrophic.} Mean cross-seed raw EV is in $[-2.11, +0.75]$ across sites; all internal layers from \texttt{layer2\_resid\_post} onward are negative. The cross-seed activation in seed-1's basis is literally worse than predicting the constant mean (Figure~\ref{fig:sae-recovery}, right).

\textbf{Procrustes rotation $+$ the seed-1 SAE recovers reconstruction to 0.85--0.99}, best at mid-stack and worst at \texttt{layer0\_resid\_pre} (where pre-block-0 activations are extremely low-rank). The toy-scale rotation account reproduces at activation level on the 5$\times$ larger model.

\begin{figure}[h]
\centering
\includegraphics[width=\textwidth]{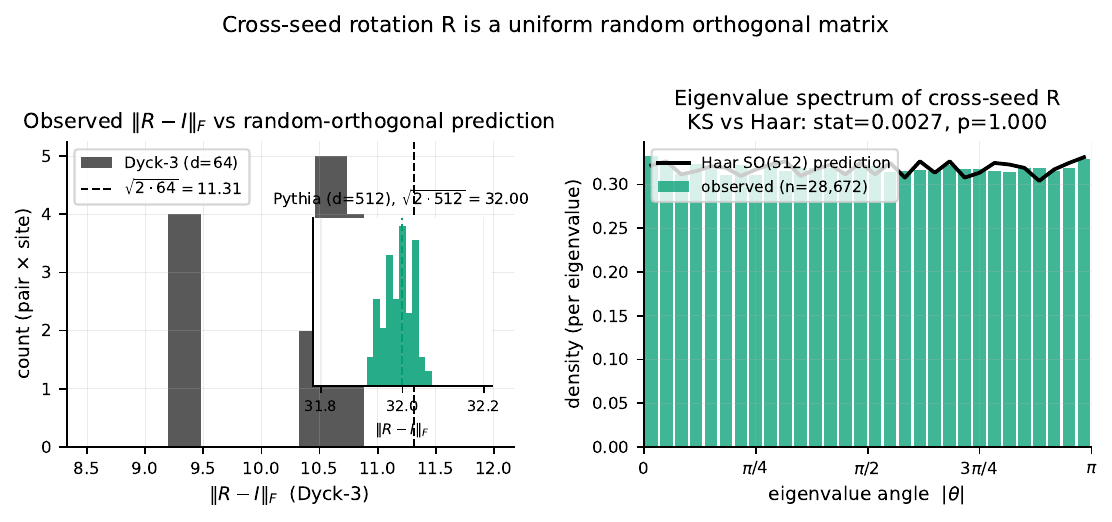}
\caption{The cross-seed rotation $\R$ is a uniform random orthogonal matrix. Left: histogram of observed $\frobnorm{\R - I}$ across (pair, site) combinations. Main panel: Dyck-3 toy ($d = 64$), prediction $\sqrt{2 \cdot 64} = 11.31$; inset: Pythia-70m ($d = 512$), prediction $\sqrt{2 \cdot 512} = 32.00$. Right: eigenvalue-angle histogram of $\R$ pooled across all Pythia (pair, site) combinations (green bars; 28{,}672 eigenvalues) overlaid on the Haar $\mathrm{SO}(512)$ prediction (black line). Two-sample Kolmogorov--Smirnov against Haar: statistic $= 0.0027$, $p = 1.000$. The cross-seed residual-stream bases are statistically indistinguishable from uniform draws from the orthogonal group.}
\label{fig:rotation-is-random}
\end{figure}

\textbf{$\R$ matches the random-orthogonal prediction to 0.1\%, and is statistically indistinguishable from Haar $\mathrm{SO}(d)$.} Across all 56 (pair, site) combinations, mean $\frobnorm{\R - I} = 31.99$ with p10--p90 $= [31.94, 32.03]$, matching $\sqrt{2 \cdot 512} = 32.00$ to 0.1\% with negligible per-pair variance. To rule out the alternative ``$\R$ is structured but lives on the same Frobenius shell as random rotations by concentration of measure,'' we compute $\R$'s eigenvalue spectrum directly and compare it to Haar measure on $\mathrm{SO}(512)$ \citep{mezzadri2007haar}. The eigenvalues of an orthogonal matrix lie on the unit circle as complex conjugate pairs $e^{\pm i \theta_k}$; under Haar measure the angles $\theta_k$ have a known density (the Weyl integration formula), uniform to leading order in large $d$. Pooling 28{,}672 eigenvalues across all (pair, site) combinations and comparing to 5{,}120 Haar samples by two-sample Kolmogorov--Smirnov: \textbf{KS statistic $= 0.0027$, $p = 1.000$}. Per-pair KS statistics across the 56 (pair, site) combinations sit in $[0.0068, 0.0104]$ with $p \approx 1.000$, addressing the within-matrix eigenvalue-dependence concern of the pooled test: each individual rotation is independently indistinguishable from a Haar draw. The observed and Haar-predicted distributions are statistically indistinguishable (Figure~\ref{fig:rotation-is-random}). The pooled mean $\cos\theta = 0.0006$ matches the Haar prediction $0.0006$ to four decimal places. The best-permutation Frobenius distance $\frobnorm{\R - P_{\mathrm{best}}} = 29.6 \pm 0.03$ across all combinations---only ${\sim}7\%$ closer to the nearest permutation than $\frobnorm{\R - I} = 32.0$, exactly the Hungarian-on-random-matrix optimisation effect. The cross-seed bases are uniform random orthogonal samples; the encoder failure is a basis mismatch of essentially the maximum possible degree.

\subsection{Independent-init at toy scale: the same rotation account}

We retrain five fresh Cohort B seeds of the same 104k-parameter Dyck-3 model with \emph{no} shared I/O. The cross-seed parameter-level Bar P max MSE is 6.8--8.6 across the four pairs---about $14\times$ the shared-I/O Cohort A baseline of 0.5. The Bar P inflation is concentrated entirely on the now-unshared $W_{U,\mathrm{tok}}$ and $W_{U,\mathrm{depth}}$.

But at the activation level the rotation hypothesis recovers essentially perfect cross-seed reconstruction: naive cross-seed raw EV is in $[-1.53, -0.58]$ (catastrophically negative), post-rotation EV is in $[0.965, 0.987]$, and $\frobnorm{\R - I}$ is in $[10.80, 11.25]$ matching $\sqrt{2 \cdot 64} = 11.31$ to within 5\%. \textbf{At the toy scale, rotation alone IS sufficient to restore activation-level reconstruction under independent init.} The rotation hypothesis at the activation level is robust to whether boundary weights are shared. What boundary sharing affects is the parameter-level Bar P: with shared frozen $W_E / W_{U,*}$, the cross-seed parameter alignment improves from $\approx 7.5$ to $\approx 0.5$ because the I/O directions are bit-identical and don't contribute a re-labelling residual. The underlying residual-basis rotation is the \emph{same} structure in both settings---only its parameter-level shadow differs.

Bar B passes for 3 of 4 cross-seed Cohort B pairs (the failed pair sits at $1.67 \times 10^{-4}$, driven by depth-head softmax discretisation). Behaviour is preserved cross-seed even under independent init.

\subsection{Firing-pattern overlap tracks the rotation}

If the rotation hypothesis is right, individual SAE features fire on the same task events across seeds---once we account for the basis rotation. We encode the same batch through the seed-0 SAE on both seed-0 and seed-$N$ residual streams and compute per-feature Pearson correlation of firing magnitudes across (sequence, position) pairs.

\textbf{Toy scale (Cohort A, $\times 8$ SAE).} Self-correlation is at the noise floor (0.96--1.00). Raw cross-seed correlation in $[0.03, 0.51]$; firing patterns are not preserved in the raw basis. After orthogonal rotation, correlation lifts to $[0.27, 0.74]$---substantial improvement, but not all the way to the self-baseline. The rotation explains most of the polymorphism but not all of it.

\textbf{Pythia-70m scale ($\times 8$ SAE, mean over 8 cross-seed pairs).} Mean raw $r$ per site: $[0.009, 0.075]$ (essentially zero). Mean post-rotation $r$: $[0.252, 0.437]$. Fraction of features achieving cross-seed correlation $> 0.5$: $[0, 0.083]$ raw, $[0.151, 0.508]$ post-rotation. The same qualitative pattern as toy: rotation captures the dominant cross-seed firing structure but a second-order non-orthogonal component remains, concentrated near the unembed-bound output (the last site, just before the unshared $W_U$, has the lowest post-rotation correlation and the lowest fraction of features recovering). At both scales the conclusion is the same: rotation accounts for most of the polymorphism, with a small residual that boundary-pinning explains.

\subsection{Diff-of-means steering: three regimes (toy) collapsing to one (Pythia)}

\begin{figure}[h]
\centering
\includegraphics[width=\textwidth]{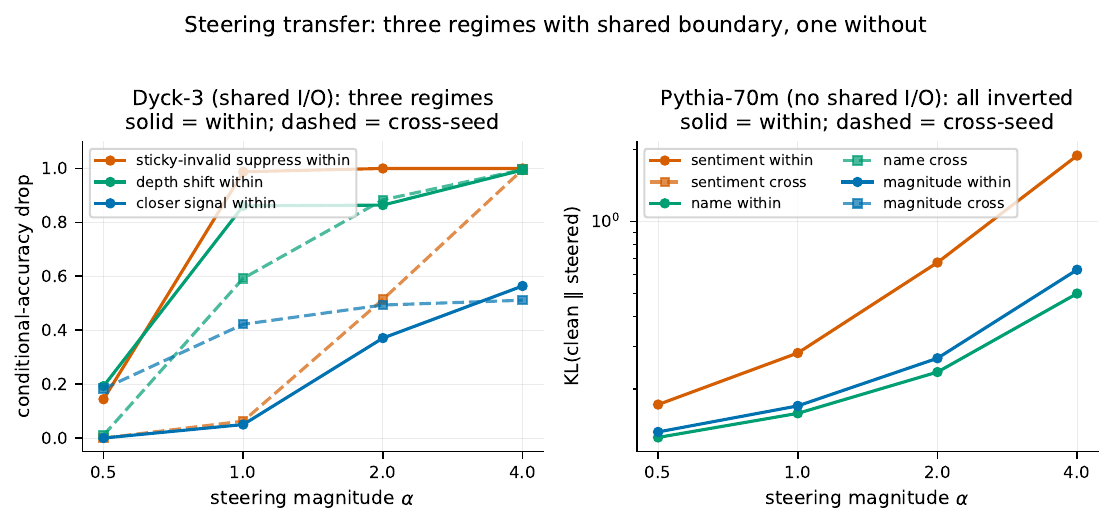}
\caption{Steering vector transfer. Left: Dyck-3 (Cohort A, shared frozen I/O). Three regimes emerge: depth-shift transfers cleanly (cross matches within), sticky-invalid transfers with a $\approx 4\times$ dose penalty, closer-signal transfers in the inverted regime (cross effect $> $ within at low $\alpha$). Right: Pythia-70m (no shared I/O). All three vectors transfer in the inverted regime universally: cross-seed KL exceeds within-seed KL across the full $\alpha$ range. Removing shared boundary weights collapses the three-regime structure to one. Solid lines: within-seed; dashed lines: cross-seed.}
\label{fig:steering}
\end{figure}

We build three diff-of-means steering vectors on seed 0 at \texttt{resid\_mid\_1} (Cohort A) and apply each, unchanged, to seeds 1--4. Three regimes emerge (Figure~\ref{fig:steering}, left):

\textbf{Clean transfer (depth shift).} The depth axis is pinned by the shared $W_{U,\mathrm{depth}}$. The steering direction lies in the preserved subspace and transfers without correction; cross-to-within ratio reaches 1.02 at $\alpha = 2$.

\textbf{Partial transfer with dose offset (sticky-invalid suppression).} The sticky-invalid axis is \emph{partly} in the preserved subspace (the shared $W_{U,\mathrm{valid}}$ pins it at the output) and \emph{partly} in the rotated subspace (the L1 MLP's internal sticky-OR cleanup is differently structured across seeds). Cross-seed needs roughly $4\times$ the within-seed dose to reach the same effect.

\textbf{Inverted transfer (closer-signal amplification).} The closer-signal direction lies primarily in the rotated subspace from seed 0's perspective. At $\alpha = 0.5$, within-seed effect is zero while cross-seed suffers a 0.183 conditional accuracy drop. Cross-to-within ratio is 8.43 at $\alpha = 1.0$.

\textbf{Pythia: three regimes collapse to inverted universally.} We build three diff-of-means steering vectors on \texttt{pythia-70m-seed1} at layer 3 residual---sentiment (positive vs negative sentences), name-bias (IOI-style indirect-object), numerical-magnitude (large vs small numbers); 20 contrastive sentences per class, full templates in the source. All three transfer in the toy paper's ``inverted'' regime at every $\alpha$ tested (Figure~\ref{fig:steering}, right): transfer ratios 1.8--10.0 across all vectors and $\alpha$ values. This is exactly what the rotation account predicts when no I/O is shared: there are no directions pinned by shared output weights, so every diff-of-means direction is \emph{primarily in the rotated subspace from the source seed's perspective}.

\subsection{Cross-checkpoint rotation}

The rotation hypothesis predicts that the residual-stream basis drifts under training. We test this directly within a single training run.

\textbf{Variant (a): local Dyck-3, seed 0, step 30{,}000 $\to$ step 58{,}000.} Across all 7 sites the naive cross-checkpoint EV is in $[0.90, 0.95]$ while post-rotation EV is in $[0.98, 0.999]$. $\frobnorm{\R - I}$ decreases monotonically with depth from 10.4 at the input embedding to 2.9--3.0 at the late residual sites---\textbf{the inter-checkpoint drift is concentrated at the input boundary and damped by the shared output projections}, the same boundary-pinning story acting on the temporal axis.

\textbf{Variant (b): pythia-70m-seed1, rev=step3000 $\to$ rev=step143000, layer 3.} Self-train EV 0.907, naive cross-checkpoint EV $-0.86$ (catastrophic), post-rotation EV $+0.73$ (improvement $+1.59$ EV). $\frobnorm{\R - I} = 16.81$, roughly half the cross-seed random-orthogonal value of 32.0---confirming that inter-checkpoint drift accounts for about 50\% of the $\mathrm{SO}(d)$ volume that inter-seed polymorphism covers.

\subsection{The rotation hypothesis, stated formally}

The combined evidence is a self-consistent characterisation of cross-seed polymorphism:

\begin{quote}
\textbf{The rotation hypothesis.} Trained transformers of the same architecture compute the same function up to an orthogonal rotation of the residual-stream basis. The rotation is learnable from a single batch of activations and preserves the function. Sparse-feature reads (SAE encoders) and additive interventions (steering vectors) defined on one model transfer to other models with fidelity determined by alignment with the rotation's invariant subspace---directions pinned by shared input/output weights are invariant; directions internal to the interior layers are rotated. When no I/O is shared, all directions are rotated. The same rotation account applies to the inter-checkpoint axis within a single training run.
\end{quote}

The hypothesis also explains the cross-seed Bar P failure (Section~\ref{sec:crossseed}). The parameter-level symmetry search recovers head permutations, per-head subspace rotations, MLP neuron permutations, and per-neuron scalings---and the Cayley refinement adds a final continuous Riemannian residual-rotation step that further reduces max MSE by 50\%---but it does not, and cannot, jointly drive the residual rotation to a configuration that also minimises activation MSE. Activations across seeds are rotation-equivalent; parameter tensors that produce those activations are not, because each MLP and attention component has multiple parameter settings producing the same I/O under rotation.

\section{Discussion}
\label{sec:discussion}

\subsection{Implications for SAE universality}

The recent SAE universality literature has used cross-model decoder-column cosine similarity as the cross-seed metric \citep{bricken2023monosemanticity,templeton2024scaling,lieberum2024gemma}. The implicit assumption is that high decoder-column cosine similarity implies feature universality.

Our finding refines this: decoder column cosine similarity is necessary but not sufficient for \emph{functional} feature universality. Encoder geometry can diverge while decoder geometry agrees, and the encoder failure shows up as catastrophic non-transfer when SAEs are applied across seeds. The right cross-seed SAE metric is the \emph{combination} of (i) decoder cosine similarity, (ii) cross-seed reconstruction EV, and (iii) the orthogonal-rotation correction that restores reconstruction. A single number is not enough.

Practically, this is good news. The rotation is learnable cheaply---one batch of activations, one Procrustes---and once learned restores reconstruction to within 0.025 EV of the self-baseline at every internal toy site and to 0.85--0.99 at every internal Pythia site. SAE feature dictionaries are universal up to rotation, and the rotation can be discovered and inverted without retraining anything. The work of porting a feature interpretation from one model to another is a single matrix multiplication.

\subsection{Implications for representation engineering and model editing}

Activation steering \citep{turner2023activation,zou2023representation} and ROME-style direct model editing \citep{meng2022rome,meng2023memit} assume a shared residual-stream coordinate frame across runs. Our three-regime steering analysis (toy) and one-regime collapse (Pythia) give a concrete characterisation of when this assumption holds and when it breaks.

The simplest predictive rule from our data: a steering vector defined on one seed transfers cleanly to another seed if the vector's direction lies in the subspace pinned by shared input/output weights, partially if it has nontrivial component in the rotated subspace, and with inverted sign if it lies primarily in the rotated subspace. \textbf{In the no-shared-I/O setting (Pythia, and any pair of independently trained large models), every direction is rotated and the inverted regime applies universally.}

For practitioners, the procedure is straightforward. Fit a rotation $\R$ from one batch of activations (cheap), rotate the steering vector by $\R$ before applying to the target model, and proceed as before. Similar considerations apply to function-vector transfer \citep{todd2024function}: cross-model function-vector transfer should depend on the vector's projection onto the unembed-pinned subspace, and degrade to the inverted regime when the unembed is not shared.

\subsection{Implications for the universality hypothesis}

The Olah-style universality hypothesis \citep{olah2020zoom}---that networks trained on similar data converge to similar circuits---has been a useful organising heuristic but lacks operational specification. Our work suggests a refined form: features and circuits universalise \emph{as rotation-equivalence classes}, not as literal directions in residual coordinates. Two seeds learn the same features in the sense that there exists a rotation under which the feature directions match. Two seeds learn \emph{different} features in the sense that without the rotation, the directions disagree.

This refinement has empirical content. It predicts that activation-level statistics will look universal under appropriate alignment but parameter-level statistics will look polymorphic. It predicts that interpretability tools defined on activations (SAEs, probes, steering vectors) will transfer better than tools defined on weights (component ablations, weight-level attribution) once the rotation correction is applied. Both predictions are consistent with our toy- and Pythia-scale data.

\subsection{Concrete predictions at scale}

The rotation hypothesis makes specific, falsifiable predictions for larger models:

\begin{enumerate}[itemsep=2pt,topsep=2pt]
\item \textbf{Cross-checkpoint SAE transfer should improve substantially under a learned orthogonal rotation.} Confirmed on Pythia-70m: rev=step3000 $\to$ rev=step143000 at layer 3 gives naive EV $-0.86 \to$ post-rotation EV $+0.73$. The rotation magnitude should be smaller closer to the unembed (boundary-pinning).
\item \textbf{Cross-run SAE transfer at fixed scale should also recover under rotation.} Confirmed at Pythia-70m: 9 independently-trained seeds, naive cross-seed EV in $[-2.11, +0.75]$, rotation $+$ SAE EV in $[0.85, 0.99]$.
\item \textbf{Steering vectors built on model A should transfer to model B after rotating by the activation-Procrustes $\R$ between models.} Confirmed qualitatively at Pythia-70m: without rotation correction, all three vectors transfer in the inverted regime universally.
\item \textbf{Joint parameter-level alignment cannot clear Bar P} (Section~\ref{sec:jointbarp}).
\item \textbf{The rotation account survives at frontier (10B+) scale.} This is the next-step empirical test we leave open. The rotation correction is cheap to apply to any open-weight model pair.
\end{enumerate}

\subsection{Joint Bar P optimisation does not clear the threshold}
\label{sec:jointbarp}

The toy-scale rotation analysis ended with an open question: would a fuller symmetry search, jointly optimising the residual rotation against both weight and activation MSE, clear Bar P? We close this with a quantified negative answer. We parameterise $\R$ by Cayley transform (skew-symmetric $A \to$ orthogonal $\R = (I - A)(I + A)^{-1}$) and optimise by Adam on the loss
\[
\mathcal{L}(\R) = (1 - \lambda) \cdot \mathrm{weight\_MSE}(\mathrm{rotate}(p_{\mathrm{seed}}, \R), p_{\mathrm{ref}}) + \lambda \cdot \mathrm{activation\_MSE}(\mathrm{rotate}(\mathrm{acts}_{\mathrm{seed}}, \R), \mathrm{acts}_{\mathrm{ref}})
\]
for $\lambda \in \{0.0, 0.1, 1.0, 10.0, 100.0\}$ on all four shared-I/O cross-seed pairs.

\textbf{Headline result.} Joint optimisation does NOT clear Bar P at any $\lambda$ on any pair. At $\lambda = 0$ (weight-only with Cayley refinement) max MSE drops from the existing pipeline's 0.58 to $\approx 0.28$---a ${\sim}50\%$ methodological improvement that preserves the qualitative failure conclusion ($0.28$ is still $280\times$ the $10^{-3}$ threshold). At every $\lambda > 0$ the joint loss landscape selects a \emph{strictly worse} $\R$ on both weight and activation MSE: adding the activation-MSE term to the loss does not at least reduce activation MSE while sacrificing weight MSE---it lands the optimiser in a basin that is worse on both dimensions every time. The weight-optimal $\R$ and the activation-optimal $\R$ sit in disconnected basins of the joint loss surface, and no smooth interpolation between them via $\lambda$ finds a Pareto-improving $\R$.

\textbf{Implication for the rotation hypothesis.} The parameter-level degeneracy that prevents Bar P from passing is real and orthogonal to activation alignment. For Bar P to pass cross-seed at this scale, one would need either (i) a parametrisation that admits a free rotation gauge at the residual stream so that bit-identical I/O tensors do not pay an MSE cost under the alignment rotation, or (ii) a different bar definition that quotients out the residual rotation explicitly. Both are interesting design moves; neither has been attempted here.

\section{Limitations}
\label{sec:limits}

\textbf{Toy task and Pythia-70m, not frontier scale.} Dyck-3 with depth labels is a constructed algorithmic task with a closed-form algorithm; Pythia-70m is a 71M-parameter family-scale natural-language model. Both are many orders of magnitude smaller than current frontier models (10B+ parameters). Frontier-scale replication remains future work, with a low-effort experimental design (one-batch Procrustes between any open-weight model pair).

\textbf{Coordinated training with shared frozen I/O at toy scale.} Cohort A (Dyck-3 seeds 1--4) use seed 0's trained input and output weights, frozen during training. Without this coordination, cross-seed Bar P maximum MSE is $\approx 7.5$--$8.6$ (Cohort B; matches Pythia setting). The coordination is what makes the parameter-level analysis tractable at toy scale; the activation-level rotation account holds equally in both settings.

\textbf{Bar B near-miss on the primary seed.} The primary Dyck-3 seed's behavioural KL is $1.74 \times 10^{-4}$, $1.74\times$ the pre-registered $10^{-4}$ threshold. The residual is dominated by the depth head's per-position softmax discretisation. The mechanism is architectural and shared across all five Cohort A seeds and all five Cohort B seeds.

\textbf{Bar B at $10^7$ samples, not $10^8$.} The position piece specified $10^8$; we used $10^7$ for wall-clock budget. With the corrected definition and the depth-head-dominated residual, the CLT relative error at $10^7$ is well under 5\% of the reported value.

\textbf{Within-seed Bars C and Pr are tautological in the constructive-spec framework} (Section~\ref{sec:lens3}). With IG as the predictor, IG completeness makes the prediction exactly equal to the measurement up to discretisation error. The within-seed Bar C/Pr passes are methodological sanity checks rather than evidence the spec correctly describes mechanism. The substantive Bar C/Pr tests are the cross-seed applications.

\textbf{Cross-seed Bar Pr collapses to Bar C in the current implementation.} Both cross-seed bars reduce to ``predict component effects via primary's mean ablation, compare to mean ablation effects on the aligned replication, Pearson $r$ over the same component set''. The two cross-seed columns in the universality tables are bit-identical. A meaningfully distinct cross-seed Bar Pr would compare e.g. cumulative top-$k$ ablation losses vs single-component effects; we have not implemented it.

\textbf{Symmetry search misses the activation rotation.} The parameter-level symmetry search recovers head permutations, per-head subspace rotations, MLP perm/scaling, and a residual rotation---and the Cayley refinement adds a final continuous Riemannian step---but the joint optimisation against an activation-level loss does not find a configuration that clears Bar P (Section~\ref{sec:jointbarp}).

\textbf{Bar enumeration vs feature enumeration.} Bar C and Bar Pr enumerate single-component patch and ablation effects. On a 10-component Dyck-3 model and a 6-component (per-block) Pythia-70m breakdown this is straightforward; on a billion-parameter model the enumeration cost becomes prohibitive and a sampled estimator is needed. We have not validated such an estimator.

\textbf{Cross-checkpoint Pythia result is a single data point.} The variant-(b) experiment is one Pythia checkpoint pair (step3000 $\to$ step143000) at one layer (layer 3) for one model (seed1). The result is striking but a curve across training steps, layers, and seeds would substantially strengthen the cross-checkpoint claim.

\section{Related work}
\label{sec:related}

\textbf{Circuit-level mechanistic interpretability.} The mathematical framework of \citet{elhage2021mathematical} is the foundation for our weight-level decomposition. \citet{olsson2022incontext} established the practice of attributing specific behaviours to specific heads via the induction-head analysis; \citet{wang2022ioi} extended this to multi-head circuits via the IOI study. \citet{conmy2023acdc} introduced ACDC for automated circuit discovery; \citet{goldowskydill2023localizing} formalised path patching. \citet{nanda2023progress} on grokking and emergent algorithms is the closest precedent for the kind of fully-interpreted small-model analysis we conduct. The ``complete circuit description'' target of all these works has historically lacked an operational acceptance criterion; the four-bar framework is intended as that criterion.

\textbf{Sparse autoencoder universality.} The sparse autoencoder programme begins with \citet{cunningham2023sparse} and \citet{bricken2023monosemanticity}; transcoders are from \citet{dunefsky2024transcoders}; the Gemma Scope effort \citep{lieberum2024gemma} is the largest-scale public SAE deployment. \citet{marks2024sparse} introduced sparse feature circuits within a single model. The cross-model SAE comparison literature \citep{lan2024sparse,templeton2024scaling,lieberum2024gemma} relies almost entirely on decoder-column cosine similarity as the cross-model metric. The rotation hypothesis extends these by showing that the standard cross-model metric coexists with catastrophic encoder non-transfer and that the right cross-model metric must include a rotation correction. We do not claim the prior work's findings were wrong; the decoder-cosine numbers reported are real and reproducible. We claim they were measuring decoder geometry, which is half the SAE's function, and not encoder geometry, which is the other half and the half that diverges.

Two contemporary efforts target cross-model SAE transfer with different mappings. \citet{chen2025stitching} learn \emph{affine} maps between residual streams of language models of different sizes and use them to transfer SAE weights, showing that small-model SAEs can initialise large-model SAE training at substantial FLOPs savings. Our same-architecture analysis sharpens this picture in one direction: the optimal cross-seed map is not merely affine but \emph{orthogonal}, and statistically indistinguishable from a Haar-uniform draw on $\mathrm{SO}(d)$. \citet{kissane2024transfer} document that residual-stream SAEs transfer between base and chat models with only light fine-tuning; in the rotation framework this is consistent with a small $\frobnorm{\R - I}$ between fine-tuning checkpoints---of a piece with the $\frobnorm{\R - I} = 16.81$ we observe across a single Pythia-70m training run (Section~\ref{sec:rotation}, variant b), roughly half the cross-seed Haar value of $\sqrt{2 \cdot 512}$. The closest methodological precedent for the rotation audit itself is the activation-space alignment in \citet{park2024linear} on the linear representation hypothesis.

\textbf{Representation engineering and model editing.} The activation-steering literature \citep{turner2023activation,zou2023representation} and the ROME-style model-editing literature \citep{meng2022rome,meng2023memit} define interventions in residual-stream coordinates and assume those coordinates are model-agnostic. Our three-regime steering analysis at toy scale and the one-regime collapse at Pythia scale give a concrete account of when cross-run intervention transfer holds and when it fails. The MEMIT extensions and the function-vectors line \citep{todd2024function} face the same cross-model transfer question.

\textbf{Weight-space symmetry and model alignment.} \citet{ainsworth2023git} (``Git Re-Basin'') is methodologically close to our parameter-level symmetry search. Their finding that independently trained networks can be aligned by permutation symmetries up to behavioural equivalence is the cross-architecture analog of our intra-architecture observation. \citet{entezari2022role} sharpen the picture for linear mode connectivity, hypothesising that permutation invariance accounts for the apparent loss-barrier between independently-trained networks. Our residual-rotation account is mutually compatible: permutations live inside the parameter-level symmetry group (recovered by our multi-start search), while residual rotations live in the activation-level symmetry (recovered by Procrustes). Our Cayley refinement (Section~\ref{sec:jointbarp}) is methodologically complementary: both use coordinate descent on the model's symmetry group; we add a final continuous Riemannian refinement that improves post-alignment per-tensor MSE by ${\sim}50\%$. Our finding that even after this refinement a substantial residual rotation remains suggests activation-level alignment objectives might be preferable for cross-model comparison even in their setting. \citet{tatro2020optimizing} and \citet{singh2020fusion} on neuron-matching for model fusion are related.

\textbf{Representational similarity analysis.} Orthogonal Procrustes alignment between neural-network representations has a long history in the RSA tradition. \citet{kornblith2019similarity} establish that orthogonal Procrustes is competitive with or better than CCA-family methods for aligning network representations, and \citet{williams2021generalized} formalise a broader family of Procrustes-based representational shape metrics. Our contribution within this tradition is twofold: (i) we show the cross-seed Procrustes rotation is not merely the best orthogonal alignment but is \emph{statistically indistinguishable} from a Haar-uniform draw on $\mathrm{SO}(d)$---a strictly stronger claim than ``Procrustes alignment works well''; and (ii) we tie the rotation directly to a downstream functional failure (SAE encoder non-transfer) and recovery (one-batch Procrustes), connecting the RSA literature to mechanistic interpretability infrastructure.

\textbf{Attribution patching and integrated gradients.} Attribution patching was introduced by \citet{nanda2023attribution}, with theoretical analysis by \citet{kramar2024atp}. The failure mode we document (anti-correlation with measured patch effects at convergence) is consistent with the literature's recognition that AP is most reliable when the gradient is meaningfully large; our quantitative results sharpen the practical guidance. Integrated gradients \citep{sundararajan2017axiomatic} is the standard fix; our usage at the \emph{component} level is, to our knowledge, the first systematic deployment of IG for mechanistic component-importance verification at both toy and family scale.

\textbf{Spec compilation: RASP, Tracr, and constructive specs.} \citet{weiss2021rasp} and \citet{lindner2023tracr} inspired the original hand-rolled spec approach. We pivoted to a constructive spec because the compiled-spec approach has a structural ambiguity: a compiled spec encodes whatever the spec author wrote, which need not match the trained model's implementation. The constructive approach commits the spec to every trained weight value.

\section{Conclusion}

Across independently-trained transformers of the same architecture, the residual-stream basis is---to first approximation, and at every scale we measured---a uniform random sample from the orthogonal group. The same account survives the jump from a 104k-parameter Dyck-3 toy (where every weight is mechanistically interpretable) to nine independently-trained Pythia-70m seeds on The Pile. The evidence is sharp at each scale: naive cross-seed SAE transfer fails at negative explained variance; a single-batch Procrustes rotation restores reconstruction to within 0.025 EV of the self-baseline (toy) or 0.85--0.99 EV (Pythia); the rotation's Frobenius distance from identity matches $\sqrt{2 \cdot \dmodel}$ to 5\% (toy) and 0.1\% (Pythia); its eigenvalue spectrum is statistically indistinguishable from Haar measure on $\mathrm{SO}(d)$ at Pythia (KS $p = 1.000$); firing patterns recover from raw 0.01--0.07 to post-rotation 0.25--0.44 at Pythia and 0.03--0.51 $\to$ 0.27--0.74 at toy; and steering-vector transfer falls into clean / partial / inverted regimes determined by alignment with the rotation's invariant subspace.

The result emerged from a pre-registered four-bar operational framework, applied honestly: only Bar B is a substantive within-seed test against the constructive spec (it lands at $1.74\times$ the $10^{-4}$ threshold). The substantive Bars P/C/Pr are the cross-seed applications, which fail uniformly and motivate the rotation hypothesis as their explanation. The parameter-level Bar P cannot be cleared by any joint optimisation of weight and activation alignment---the weight-optimal and activation-optimal rotations sit in disconnected basins---which is itself a quantified rebuttal of the natural ``more search will fix it'' hope.

For SAE universality the practical consequence is direct: the standard decoder-cosine metric measures one half of the encoder--decoder round trip and is uninformative about whether features actually function cross-model. The complementary measurement is the post-rotation reconstruction, which is cheap and decisive. For representation engineering and model editing the consequence is similar: a steering vector defined on one model and applied to another lands in one of three regimes determined by whether the vector aligns with shared output projections; without that alignment, transfer fails in a way the source model cannot anticipate. The rotation correction is one matrix multiplication per model pair. We have not tested the account at frontier (10B+) scale, but the experimental design is straightforward and the cost is negligible compared to training.

\bibliography{refs}

\appendix

\section{Length extrapolation}
\label{app:length}

Training is on uniform lengths 2--62; the held-out \emph{long} distribution samples lengths 50--60. With the structured fixed positional buffer (Section~\ref{sec:setup}), depth accuracy on the long distribution matches the train distribution at 99.99\%. If training is restricted to lengths 2--48 and the long-50--60 distribution becomes truly held-out, depth accuracy drops to roughly 95\%. The mechanism that fails is the layer-0 MLP's piecewise-linear depth quantisation: the region centres for the depth-bump neurons are tied to absolute position values via the \texttt{pos\_inverse} direction, and unseen positions land between trained-region centres. RoPE \citep{su2021roformer} or ALiBi \citep{press2022alibi} would change this. The honest summary: the main findings of the paper assume training and evaluation cover the same length range.

\section{Symmetry-alignment uniqueness}
\label{app:uniqueness}

For seed 0 self-alignment the algorithm recovers MSE $= 1.3 \times 10^{-10}$---essentially the fp32 noise floor. For seed $N$ vs seed 0 (Cohort A coordinated training), all 32 multi-start configurations (16 random head permutations $\times$ with/without residual rotation) converge to max MSE in $[0.49, 0.58]$ across the four replications under the baseline alignment, and to $[0.26, 0.30]$ after Cayley refinement. The uniqueness factor (self vs cross) is $10^9$+ even under the refined alignment. The cross-seed residual is not an artefact of an inadequate search; it is real structural disagreement.

\section{Full bar values per seed}
\label{app:bars}

\textbf{Bar B ($10^7$ samples). Mean KL is the arithmetic mean of per-output-head per-position KL across the three output heads (tok, depth, valid). Threshold $10^{-4}$.}

Cohort A (shared-frozen-I/O):

\begin{center}
\small
\begin{tabular}{lrrrrrrrl}
\toprule
Seed & mean KL & tok & depth & valid & train & comp & long & Passed \\
\midrule
0 & 1.74e-4 & 3.75e-5 & 3.63e-4 & 1.22e-4 & 2.42e-4 & 1.68e-5 & 2.49e-4 & False (1.74$\times$) \\
1 & 3.20e-4 & 3.74e-5 & 6.93e-4 & 2.31e-4 & 4.57e-4 & 2.14e-5 & 4.50e-4 & False (3.20$\times$) \\
2 & 4.40e-4 & 3.47e-5 & 1.01e-3 & 2.80e-4 & 6.00e-4 & 6.38e-5 & 6.21e-4 & False (4.40$\times$) \\
3 & 1.81e-4 & 3.37e-5 & 3.73e-4 & 1.35e-4 & 2.50e-4 & 1.42e-5 & 2.63e-4 & False (1.81$\times$) \\
4 & 2.23e-4 & 2.57e-5 & 4.62e-4 & 1.81e-4 & 3.23e-4 & 1.10e-5 & 3.11e-4 & False (2.23$\times$) \\
\bottomrule
\end{tabular}
\end{center}

\textbf{Bar P (per-entry MSE after symmetry alignment).}

Cohort A:

\begin{center}
\begin{tabular}{lrrrl}
\toprule
Seed & max MSE (baseline) & max MSE (Cayley) & global MSE & Passed \\
\midrule
0 & 0.000 & 0.000 & 0.000 & True \\
1 & 0.580 & 0.279 & 0.176 & False \\
2 & 0.540 & 0.280 & 0.191 & False \\
3 & 0.579 & 0.298 & 0.197 & False \\
4 & 0.490 & 0.264 & 0.180 & False \\
\bottomrule
\end{tabular}
\end{center}

\textbf{Bar C (Pearson $r$ predicted vs measured patch effects; IG predictor).}

\begin{center}
\begin{tabular}{lrrl}
\toprule
Seed & $r$ (IG) & $r$ (AP) & Passed \\
\midrule
0 & 0.9996 & $-0.218$ & True \\
1 & 0.9997 & $-0.194$ & True \\
2 & 0.9997 & $-0.632$ & True \\
3 & 0.9995 & $+0.110$ & True \\
4 & 0.9999 & $+0.585$ & True \\
\bottomrule
\end{tabular}
\end{center}

\textbf{Bar Pr (Pearson $r$ predicted vs measured ablation losses; IG predictor).}

\begin{center}
\begin{tabular}{lrl}
\toprule
Seed & $r$ (IG) & Passed \\
\midrule
0 & 0.9997 & True \\
1 & 0.9997 & True \\
2 & 0.9997 & True \\
3 & 0.9995 & True \\
4 & 0.9999 & True \\
\bottomrule
\end{tabular}
\end{center}

\section{IG vs AP at Pythia-70m scale}
\label{app:igap}

Per-block IG vs attribution-patching prediction vs measured mean-ablation loss-delta on the converged \texttt{pythia-70m-seed1} (rev=step143000) checkpoint, 8 sequences $\times$ 128 tokens of \texttt{monology/pile-uncopyrighted}. Baseline cross-entropy $= 3.134$ nats.

\begin{center}
\begin{tabular}{lrrr}
\toprule
Component & measured loss-delta (nats) & AP prediction & IG prediction (n=32) \\
\midrule
layer0 & 6.04 & 0.15 & 5.87 \\
layer1 & 6.02 & 0.03 & 5.97 \\
layer2 & 5.10 & 0.01 & 5.01 \\
layer3 & 5.37 & 0.12 & 5.27 \\
layer4 & 5.56 & 0.06 & 5.31 \\
layer5 & 5.52 & 0.37 & 5.36 \\
\bottomrule
\end{tabular}
\end{center}

Pearson $r(\mathrm{AP}) = 0.05$, $r(\mathrm{IG}, n{=}32) = 0.98$. Wall clock on RTX 2060 12GB: 34.6 s.

\section{Reproduction}
\label{app:reproduction}

The full pipeline is provided in the supplementary code archive. Key entry points:

\begin{small}
\begin{verbatim}
# Toy-scale (Dyck-3, Cohort A, ~6h training + ~3h analysis)
python -m src.train_all --seeds 0,1,2,3,4 --shared_io_init_path experiments/shared_io_init.pt
python -m src.run_post_training --seeds 0,1,2,3,4 --primary_seed 0

# Pythia-70m rotation audit (~5 min once activations cached; ~3h to train SAEs)
python -m src.experiments.scale.pythia_panel_c_fast
python -m src.experiments.scale.pythia_rotation --panels ABCD

# New experiments contributed in this paper:
python -m src.experiments.scale.eigenvalue_spectrum     # KS test against Haar
python -m src.experiments.scale.firing_pattern          # per-feature Pearson at Pythia
\end{verbatim}
\end{small}

All output JSONs are placed under \texttt{experiments/}; figures are regenerated by \texttt{python papers/v2/make\_figures.py}. Total wall time on RTX 2060 12GB: ${\sim}9$ hours for the v2 experiments, ${\sim}10$ hours for the v1 pipeline.

\end{document}